\documentclass{article}
\usepackage{spconf,amsmath,graphicx}
\usepackage{hyperref}
\usepackage{xcolor}
\usepackage{amssymb}
\usepackage{enumitem}
\setlist{nosep, leftmargin=14pt}

\usepackage{mwe} % to get dummy images

\title{Quantitative Gait Analysis from Single RGB Videos Using a Dual-Input Transformer-Based Network}
\name{Hiep Dinh$^{2}$ \qquad Son Le$^{1,2}$ \qquad My Than $^{3}$ \qquad  Minh Ho$^{3}$ \qquad  Nicolas Vuilerrme $^{4,5}$ \qquad  Hieu Pham$^{1,2,4,*}$\thanks{* Corresponding author: hieu.ph@vinuni.edu.vn (Hieu Pham)}}
\address{   $^{1}$ College of Engineering \& Computer Science, VinUniversity, 100000 Hanoi, Vietnam \\
            $^{2}$ VinUni-Illinois Smart Health Center, VinUniversity, 100000 Hanoi, Vietnam \\
         $^{3}$ Orthopaedics and Sports Medicine Center, Vinmec Healthcare System, 100000 Hanoi, Vietnam \\ $^{4}$ AGEIS, Université Grenoble Alpes, 38000 Grenoble, France \\ $^{5}$ Institut Universitaire de France, 75005 Paris, France}

\begin{document}
%\ninept
%
\maketitle
\begin{abstract}
Gait and movement analysis have become a well-established clinical tool for diagnosing health conditions, monitoring disease progression for a wide spectrum of diseases, and to implement and assess treatment, surgery and or rehabilitation interventions. However, quantitative motion assessment remains limited to costly motion capture systems and specialized personnel, restricting its accessibility and broader application. Recent advancements in deep neural networks have enabled quantitative movement analysis using single-camera videos, offering an accessible alternative to conventional motion capture systems. In this paper, we present an efficient approach for clinical gait analysis through a dual-pattern input convolutional Transformer network. The proposed system leverages a dual-input Transformer model to estimate essential gait parameters from single RGB videos captured by a single-view camera. The system demonstrates high accuracy in estimating critical metrics such as the gait deviation index (GDI), knee flexion angle, step length, and walking cadence, validated on a dataset of individuals with movement disorders. Notably, our approach surpasses state-of-the-art methods in various scenarios, using fewer resources and proving highly suitable for clinical application, particularly in resource-constrained environments.
\end{abstract}
\section{Introduction}
Gait analysis, considered a fundamental biomechanics assessment method, first developed at the end of the 18th century \cite{Baker2007}. It holds significant value in clinical settings worldwide, playing a crucial role in rehabilitation protocol planning, fall detection, injury prevention, sports science, and even the early detection of neurological disorders such as stroke and traumatic brain injury (TBI), as well as neurodegenerative diseases including Parkinson's disease (PD), cerebellar ataxia (CA), and multiple sclerosis (MS) \cite{Bonanno2023, Ferrarello2013, Dasgupta2021, Connor2018}.

Observation-based gait analysis was widely adopted by physicians in the early stages due to its simplicity, convenience, and low cost \cite{Wallmann2009}. However, like other qualitative methods, the reliability, validity, and specificity of this approach are questionable \cite{Toro2003, Ferrarello2013}. Moreover, evidence suggests that subjective assessments may lead to inaccurate treatment decisions for patients due to various factors, including misunderstandings of the underlying mechanisms \cite{Sinclair2014}.

\begin{figure} 
    \centering
    \includegraphics[width=1\linewidth]{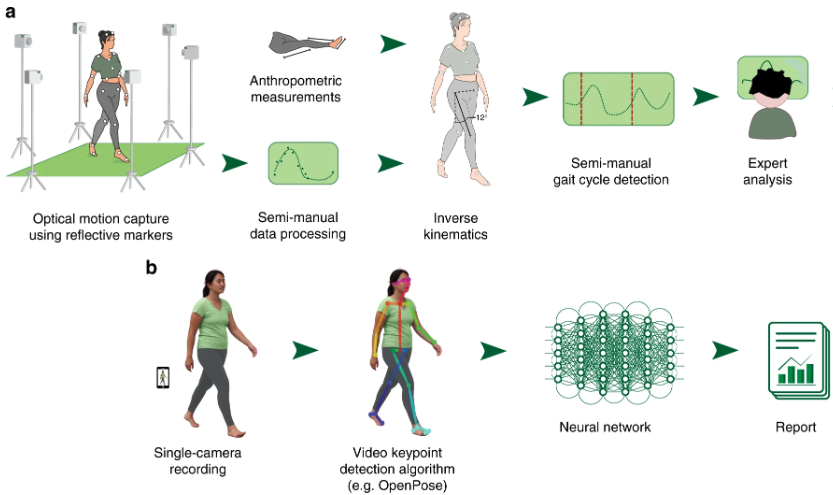}
    \caption{Motivation for the proposed approach. Rather than relying on an ultra-expensive motion capture system within the current clinical workflow, we propose in this work capturing motion data with a single standard mobile camera. Using the OpenPose algorithm \cite{cao2017realtime}, keypoint trajectories are extracted from sagittal-plane video, then connected and fed into a dual-input Transformer network \cite{vaswani2017attention} to derive clinically relevant metrics such as GDI, knee flexion angle, step length, and walking cadence. The figure is reused from Kidzinski \textit{et al.} \cite{kidzinski2020deep} for illustration purpose.}
    \label{fig:enter-label}
\end{figure}

The first approach for gait assessment, based on spatial and kinematic parameters—and considered the gold standard—is the optical motion capture system \cite{Cappozzo1984}. However, this system is not easily accessible for several reasons, such as the high cost of infrared cameras, the detailed requirements to set up a motion lab with strict standards, and the need for highly trained personnel to place reflective markers as well as operate the complex system \cite{Song2023}. Therefore, despite its exceptional accuracy, the mocap system is still considered impractical and difficult to apply widely due to its resource-intensive nature and the fact that the accuracy of results heavily depends on the quality of personnel \cite{Sinclair2014}.

New advances in computer vision and deep learning have introduced new possibilities for quantitative gait analysis, allowing for markerless methods using standard video footage \cite{stenum_rossi_roemmich_2021,DBLP:journals/corr/abs-1801-07455,DBLP:journals/corr/abs-2012-06399}. Some recent studies \cite{kidzinski2020deep,le2024learning} showed that a deep learning model is able to learn and estimate directly critical gait parameters such as key gait parameters such as GDI, knee flexion angle, step length, and walking cadence. These approaches first used frameworks like OpenPose and DeepLabCut \cite{mathis2018deeplabcut} to estimate skeletal keypoints from two-dimensional videos without the need for specialized hardware \cite{stenum_rossi_roemmich_2021}. Then deep neural networks (DNNs) were used to detect key anatomical landmarks and predict joint positions in real-time \cite{halvorsen_peng_olsson_åberg_2024}. Despite their potential for lower-cost implementation, these approaches are not without limitations. The accuracy of markerless systems can be impacted by variations in camera angles, lighting conditions, and occlusions in the video, leading to less precise measurements compared to Mocap systems \cite{stenum_rossi_roemmich_2021}. Additionally, pose estimation algorithms can struggle with robustness and consistency, especially when applied to patients with severe gait abnormalities \cite{halvorsen_peng_olsson_åberg_2024}. Moreover, current state-of-the-art methods such as \cite{kidzinski2020deep} used multiple deep networks to estimate gait parameters. This raised a need in developing single, more robust and efficient deep learning models for this task. 

This paper introduces a more efficient and accessible solution for clinical gait analysis. This novel framework leverages a dual-pattern input convolutional transformer network, to capture both spatial and temporal dynamics of gait patterns using standard video data. The self-attention mechanism enables the model to focus on relevant anatomical keypoints and their relationships across time, improving the accuracy of gait parameter predictions. By reducing the reliance on expensive equipment and eliminating the need for manual feature extraction, this method offers a cost-effective alternative that can be implemented in diverse clinical environments.

Through rigorous experiments on a dataset from individuals with movement-related disorders, the proposed approach demonstrated superior performance in predicting key gait parameters such as GDI, knee flexion angle, step length, and walking cadence. The results indicate that the proposed model not only matches but in some cases surpasses the accuracy of state-of-the-art methods, all while requiring fewer resources for model training. Our main contributions can be summarized as follows:
\begin{itemize}
\item We present a dual-input Transformer network designed to quantitatively estimate critical gait parameters from single-view RGB videos. This architecture enables more effective extraction and learning of motion features from anatomical keypoints captured in single-camera RGB footage.\\

\item The proposed network is easy to train and able to help to achieve superior performance in predicting gait parameters compared to current state-of-the-art methods.\\

\item  To encourage the development of the field, we released our codes and trained models. All are \textcolor{blue}{\href{https://github.com/lmtszrl/Quantitative-Gait-Analysis-From-RGB-Videos-Using-a-Dual-Input-Transformer-Based-Network}{made publicly available}}.\\
\end{itemize}
The rest of the paper is organized as follows: Section \ref{sect2} presents an overview of the formulation for gait estimation from RGB videos and describes in details our proposed approach. In Section \ref{sect3}, we report and discuss the experimental results. Finally, in Section \ref{sect4}, we conclude the paper and suggest potential directions for future work.

\section{Methodology}
\label{sect2}
\begin{figure*}[ht]
    \centering
    \includegraphics[width=\textwidth]{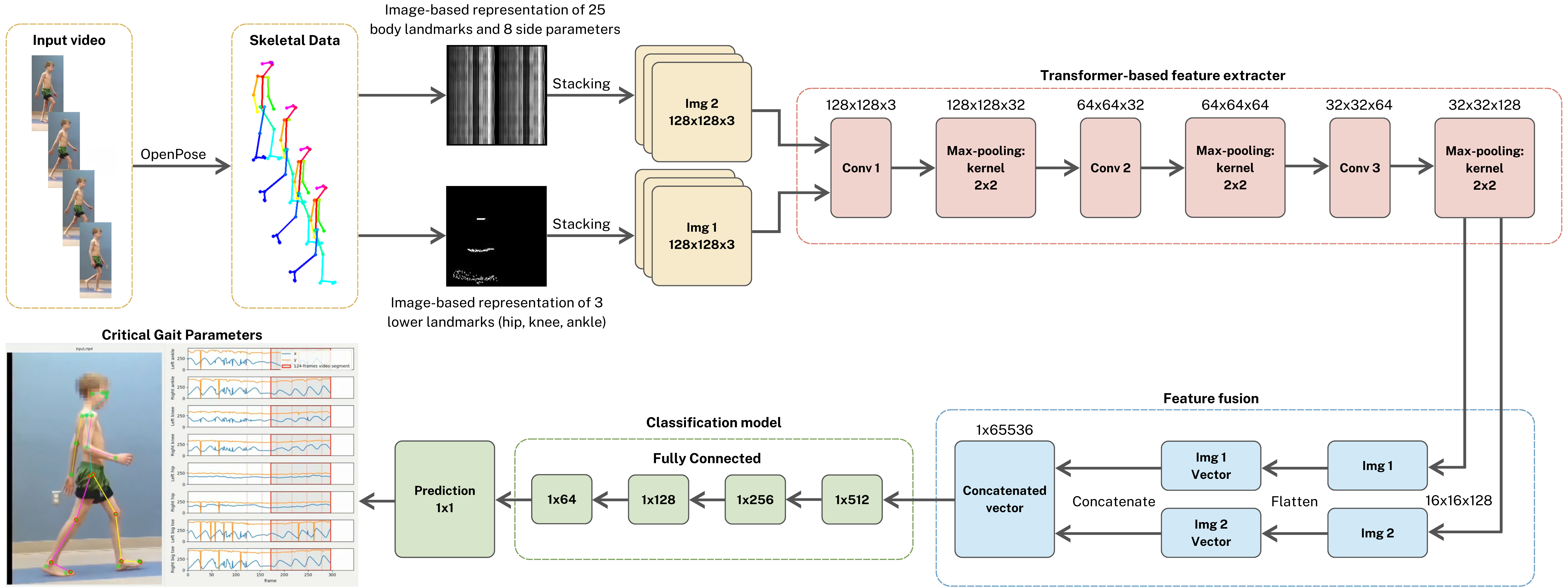}
    \caption{An overview of our approached DPG model. First, it is designed to process skeletal sequence data extracted from original videos using OpenPose. Each skeletal sequence is represented by two images: a lower-body landmark coordination pattern image and a lower-body landmark coordination image. The model architecture consists of a primary 3-layer convolutional block, each layer containing a convolution and max-pooling component, to extract spatial features. Finally, a fully connected network with four layers successively reduces the data vector size from 65536 to 512, 256, 128, and 64, outputting a single prediction vector of size 1$\times$1.}
    \label{fig:overview_approach}
\end{figure*}

\subsection{Problem Formulation}
The task is described as receiving an input sequence of motion data \(\textbf{\textit{X}} \in \mathbb{R}^{T \times N \times 2}\), where \(\textbf{\textit{X}} = [x_1, x_2, \ldots, x_T]\), and \(T\) represents the total number of frames in a video clip. Each frame \(x_i\) captures a human posture at time step \(i\), involving \(N\) joints, with each joint \(j_i^{(n)}\) represented by a 2D coordinate \((x, y)\) in the Cartesian plane. The output \(y \in \mathbb{R}^1\) signifies a specific gait metric for the patient (i.e., GDI, knee flexion angle, step length, and walking cadence). 

We formulate this task as a supervised learning challenge where the goal is to learn a non-linear mapping \(f_\theta: \mathbb{R}^{T \times N \times 2} \rightarrow \mathbb{R}^1\), parameterized by \(\theta\). Given a dataset \(D = \{(\textbf{\textit{X}}_k, y_k) : \textbf{\textit{X}}_k \in \mathbb{R}^{T \times N \times 2}, y_k \in \mathbb{R}^1, k = 1, 2, \ldots, S\}\), which includes \(S\) samples, the objective is to determine the best parameters \(\theta^*\) by minimizing a loss function \(L\) over the entire dataset \(D\). The optimization problem is expressed as

\[
\theta^* = \arg \min_{\theta} \sum_{k=1}^{S} L(f_\theta(\textbf{\textit{X}}_k), y_k).
\]

\subsection{Network Architecture}
We describe in this section a simple CNN-based transformer model with input as 2 sets of patterned images, which called Dual-Pattern Gait model (DPG) as following. The proposed model is a convolution neural network (CNN) designed to predict a gait-related parameter based on two input images, such as frames of the left and right legs. The architecture processes each input image independently through a series of three convolution layers. Each input image \( \textbf{\textit{X}} \in \mathbb{R}^{128 \times 128 \times 3} \) is passed through the convolution layers, where the first convolution layer applies 32 filters of size \( 3 \times 3 \) with stride 1 and padding 1, generating an intermediate feature map \( \textbf{\textit{F}}_1 \in \mathbb{R}^{128 \times 128 \times 32} \). This is followed by a ReLU activation and a max-pooling operation with a \( 2 \times 2 \) kernel, reducing the spatial dimensions by half. The same process is repeated in the second and third convolution layers, increasing the feature maps to 64 and 128 channels, respectively. The feature extraction for each image can be summarized as follows
\[
\textbf{\textit{F}}_{i+1} = \text{MaxPool}(\text{ReLU}(\text{Conv}(\textbf{\textit{F}}_i))),
\]
where \( \textbf{\textit{F}}_0 \) represents the input image and \( i \in \{1, 2, 3\} \). After the convolution layers, the extracted feature maps \( \textbf{\textit{F}}_3 \in \mathbb{R}^{16 \times 16 \times 128} \) are flattened to produce a feature vector. Let \( \text{Flatten}(\textbf{\textit{F}}_3) \in \mathbb{R}^{32,768} \) represent the flattened output. The features from the two input images are then concatenated into a single feature vector
\[
\text{Concat} = [\text{Flatten}(\textbf{\textit{F}}_{3,1}), \text{Flatten}(\textbf{\textit{F}}_{3,2})] \in \mathbb{R}^{65,536}.
\]
This concatenated feature vector is passed through fully connected layers for regression. Each fully connected layer applies a linear transformation followed by ReLU activation and dropout for regularization. The first fully connected layer transforms the 65,536-dimensional feature vector into 512 units
\[
h_1 = \text{Dropout}(\text{ReLU}(\textbf{\textit{W}}_1 \cdot \text{Concat} + \textbf{\textit{b}}_1)) \in \mathbb{R}^{512}.
\]
Subsequent fully connected layers reduce the dimensionality stepwise, passing through 256, 128, and 64 units, before the final layer produces a single scalar output
\[
y = \textbf{\textit{W}}_{\text{out}} \cdot h_4 + \textbf{\textit{b}}_{\text{out}},
\]
where \( y \in \mathbb{R} \) represents the predicted gait parameter. To improve convergence, Xavier initialization is applied to the weights of the fully connected layers. The architecture effectively captures spatial features from each input image, concatenates the relevant information, and processes it through fully connected layers to predict the gait parameter.

\section{Experiments}
\label{sect3}
\subsection{Datasets and Experiment Settings}
To validate the proposed approach, we utilized a publicly accessible dataset from Gillette Children’s Specialty Healthcare, which was gathered from 1994 to 2015. As described in \cite{kidzinski2020deep}, this dataset comprises 2,212 video recordings featuring 1,138 individuals diagnosed with cerebral palsy. These patients have an average age of 13 years (standard deviation: 6.9 years), an average height of 141 cm (standard deviation: 19 cm), and an average weight of 41 kg (standard deviation: 17 kg). The original videos, initially recorded at a resolution of 1280 $\times$ 960, were downscaled to 640 $\times$ 480 to align with the specifications of most low-cost cameras. Each video was segmented into multiple 124-frame clips using window slicing with a 31-frame overlap. For experimental purposes, the dataset was divided into training, validation, and test sets in an 8:1:1 ratio, with no patient appearing in multiple sets. The training set includes 1,768 videos of 920 patients, the validation set includes 212 videos of 106 patients, and the test set contains 232 videos of 112 patients.

Each pre-processed video returned a 124-frame set of 25 body landmark X-Y coordinates and 8 side-parameters. First, we consider each frame of video as a row in a 128 $\times$ 64 pixel matrix. After normalized 124 (1 $\times$ 58) coordinate rows to 0-255 gray scale, we filled them on the middle of 128 $\times$ 64 matrix, then duplicated the matrix to 128 $\times$ 128 to get the square patterned images as showed in Figure 2. Second, we only get 3 landmarks of the set, included hip, knee and ankle. We plot all 3 landmarks of 124 frames on a same 128 $\times$ 128 pixel graph, which was extracted as an image represented for 3 landmarks on video. During the pre-processing step, two data sets are prepared as inputs for the Transformer network: one set includes all body landmarks across the 124-frame video, while the other includes only three leg landmarks. Both sets are fed simultaneously into the model for training.

The proposed DPG model has three convolution layers and four fully connected layers. The model training and testing procedure involved a comprehensive pipeline designed to evaluate multiple gait analysis targets, including 'KneeFlex maxExtension', 'cadence', 'steplen' and 'GDI', for both the left (‘L’) and right (‘R’) sides. The dataset was preprocessed by resizing images to 128 $\times$ 128 pixels and converting them to grayscale with three output channels, then transformed into tensors. The entire data was divided into training, validation, and test sets with the corresponding ratio of 8:1:1, to ensure that there was no overlap between the patients. For each combination on the target side, specific directories were created to store outputs and logs, which were cleared before training. A log file was initialized for each combination. Data were loaded using a custom loader function, and a model instance was created and moved to the appropriate device (GPU or CPU). The Mean Squared Error (MSE) loss function and Adam optimizer were used, with a learning rate scheduler to adjust the learning rate based on validation loss. The model was trained for a specified number of epochs with early stopping to prevent overfitting, and the best model was saved. After training, the best model was evaluated on the test set, with performance measured using the MSE loss function and results recorded in the respective directories. We used a batch size of 32 and a learning rate of 0.001. The model was trained for 30 epochs.

To quantify the effectiveness of the proposed approach, we compared the DPG's performance with the current state-of-the-art models \cite{kidzinski2020deep} on the test set. The performance of the proposed method is measured using the mean absolute error (MAE). 

\subsection{Experiment Results}

To evaluate the effectiveness of the proposed model, we compared its performance with two state-of-the-art models, 1D-CNN \cite{kidzinski2020deep} and STT \cite{le2024learning}, on the test set. The performance of the proposed model was evaluated using the mean absolute error (MAE) metrics. As shown in Table 1, the proposed dual input Transformer model achieved MAE of \textbf{5.6450}, \textbf{5.1203}, and \textbf{0.1418} with the ground truth for GDI, knee flexion, and cadence, respectively. Our method reduced MAE by 10.6\%  and 12.0\% for GDI and knee flexion, respectively. However, experimental results show that our model struggles with cadence predictions with an MAE of \textbf{0.1418}, higher than the MAEs reported by 1D-CNN \cite{kidzinski2020deep} and STT \cite{le2024learning}. 

Through thought experiments and comparisons, we observed that the 1D-CNN model \cite{kidzinski2020deep} captures only the temporal dependencies of individual joints in isolation, limiting its ability to extract spatial information between joints and necessitating reliance on hand-crafted features. In contrast, Transformer-based models like the STT \cite{le2024learning} and our proposed DPG approach can leverage long-term dependencies in spatial-temporal data, enabling greater flexibility in capturing motion features from single-camera video. This capability aids in generating gait parameters from anatomical keypoints. However, our model still lacks the functionality to detect gait cycles and strides, which are essential for determining cadence.

\begin{table}[h]
\small{
\caption{Quantitative assessment of the proposed model in
estimating gait parameters from the test set. Best results are in bold.\\}
\centering
\begin{tabular}{|c|c|c|c|c|c|c|}
\hline
\textbf{Parameters} & \multicolumn{3}{|c|}{\textbf{Mean Absolute Error}} \\ \hline
 & \textbf{1D-CNN} \cite{kidzinski2020deep} & \textbf{STT} \cite{le2024learning}  & \textbf{DPG (ours)}\\ \hline
GDI & 6.5469 & 6.3137 &  \textbf{5.6450} ($\uparrow$ 10.6\%) \\ \hline
Knee Flexion & 5.9129 & 5.8220 &  \textbf{5.1203} ($\uparrow$ 12.0\%) \\ \hline
Cadence  & \textbf{0.1035} & 0.1078 &  0.1418 \\ \hline
\end{tabular} 
\label{tab:gait_parameters}}
\end{table}

\section{Discussion and Conclusion}
\label{sect4}
In this paper, we introduced a novel approach for clinical gait analysis using a dual-pattern input convolutional transformer network. Our method employs a dual-input Transformer architecture to estimate essential gait parameters from single-view RGB video. Experimental results showed high accuracy in estimating critical metrics, including the Gait Deviation Index (GDI) and knee flexion angle, validated on a dataset of individuals with movement disorders. By encoding the video sequence as two images, our approach reduces model complexity and training/testing time. For future work, we plan to validate this method on larger, real-world datasets, envisioning a future where accessible telehealth solutions make gait analysis widely available, eliminating the need for costly lab visits. We believe this approach could play a vital role in early diagnosis and treatment of motor-related diseases, offering hope to many while reducing the burden of late-stage care on the healthcare system.	

%\section{Acknowledgements}
%This research did not receive any specific grant from funding agencies in the public, commercial, or not-for-profit sectors. The authors declare no conflicts of interest, financial or otherwise.
\section{Acknowledgment}
The research was supported by the VinUni-Illinois Smart Health Center and the French National Research Agency (ANR) in the framework of the Investissements d’avenir program (ANR-10-AIRT-05 and ANR-15-IDEX-02), and the MIAI @ Grenoble Alpes (ANR-19-P3IA-0003). This work also forms part of a broader translational and interdisciplinary GaitAlps research program.
\section{Compliance With Ethical Standards}
This study utilized publicly available human subject data from open access sources, as referenced in \cite{kidzinski2020deep}. As the data was already anonymized and accessible under an open license, ethical approval was deemed unnecessary.

\bibliographystyle{IEEEtran}
\bibliography{references}
\end{document}